\documentclass[journal, letterpaper]{IEEEtran}
\IEEEoverridecommandlockouts

\usepackage[letterpaper, left=0.75in, right=0.75in, bottom=0.75in, top=0.67in]{geometry}

\usepackage{cite}

\usepackage{url}

\usepackage{amsmath,amssymb,amsfonts}
\usepackage{algorithmic}
\usepackage{graphicx}
\graphicspath{ {./images/} }

\usepackage{tabularx, booktabs}

\usepackage[caption=false, font=footnotesize]{subfig}

\usepackage{etoolbox}

\usepackage{gensymb}
\usepackage{textcomp}
\usepackage{xcolor}

\def\BibTeX{{\rm B\kern-.05em{\sc i\kern-.025em b}\kern-.08em
    T\kern-.1667em\lower.7ex\hbox{E}\kern-.125emX}}
\begin{document}

\pagenumbering{gobble}
\newgeometry{left=0.75in, right=0.75in, bottom=0.75in, top=0.95in}

\title{\LARGE \bf Initialisation of Autonomous Aircraft Visual Inspection Systems via CNN-Based Camera Pose Estimation\\
}

\author{Xueyan Oh$^{1}$, Leonard Loh$^{1}$, Shaohui Foong$^{1}$, Zhong Bao Andy Koh$^{2}$, Kow Leong Ng$^{2}$, Poh Kang Tan$^{2}$, Pei Lin Pearlin Toh$^{2}$, and U-Xuan Tan$^{1}$

\thanks{$^{1}$X. Oh, L. Loh, S. Foong and U-X. Tan are with the Pillar of Engineering Product Development, Singapore University of Technology and Design, Singapore,
        {\tt\small xueyan\_oh@mymail.sutd.edu.sg}, \tt\small uxuan\_tan@sutd.edu.sg.}
\thanks{$^{2}$Z. B. A. Koh, K. L. Ng, P. K. Tan and P. L. P. Toh are with ST Engineering Aerospace Ltd., Singapore}
}

\makeatletter
\patchcmd{\@maketitle}
  {\addvspace{0.5\baselineskip}\egroup}
  {\addvspace{-2\baselineskip}\egroup}
  {}
  {}
\makeatother

\maketitle

\begin{abstract}
General Visual Inspection is a manual inspection process regularly used to detect and localise obvious damage on the exterior of commercial aircraft. There has been increasing demand to perform this process at the boarding gate to minimize the downtime of the aircraft and automating this process is desired to reduce the reliance on human labour. This automation typically requires the first step of estimating a camera’s pose with respect to the aircraft for initialisation. However, localisation methods often require infrastructure, which can be very challenging when performed in uncontrolled outdoor environments and within the limited turnover time (approximately 2 hours) on an airport tarmac. In addition, access to commercial aircraft can be very restricted, causing development and testing of solutions to be a challenge. Hence, this paper proposes an on-site infrastructure-less initialisation method, by using the same pan-tilt-zoom camera used for the inspection task to estimate its own pose. This is achieved using a Deep Convolutional Neural Network trained with only synthetic images to regress the camera’s pose. We apply domain randomisation when generating our dataset for training our network and improve prediction accuracy by introducing a new component to an existing loss function that leverages on known aircraft geometry to relate position and orientation. Experiments are conducted and we have successfully regressed camera poses with a median error of 0.22 m and 0.73{\degree}.
\end{abstract}


\vspace{-3mm}

\section{Introduction}
Aircraft must undergo regular inspection such as General Visual Inspection (GVI), which is a widely used technique. One of GVI’s processes involves visually examining the aircraft’s exterior for obvious damage or abnormalities and provides a means for early detection of typical airframe defects \cite{R1}. This is currently performed manually by well-trained personnel and is labour intensive as well as having high error rates \cite{R1,R2}. Several studies have explored using robotic systems such as drones and mobile robots \cite{R1,R2,R3}, as well as deep learning using only visual images \cite{R4} to automate this labourious inspection task. However, these works focus on detecting or classifying defects within images, rather than localising defects with respect to a known reference point on the aircraft.
\begin{figure}[t]
\centerline{\includegraphics[width=\columnwidth]{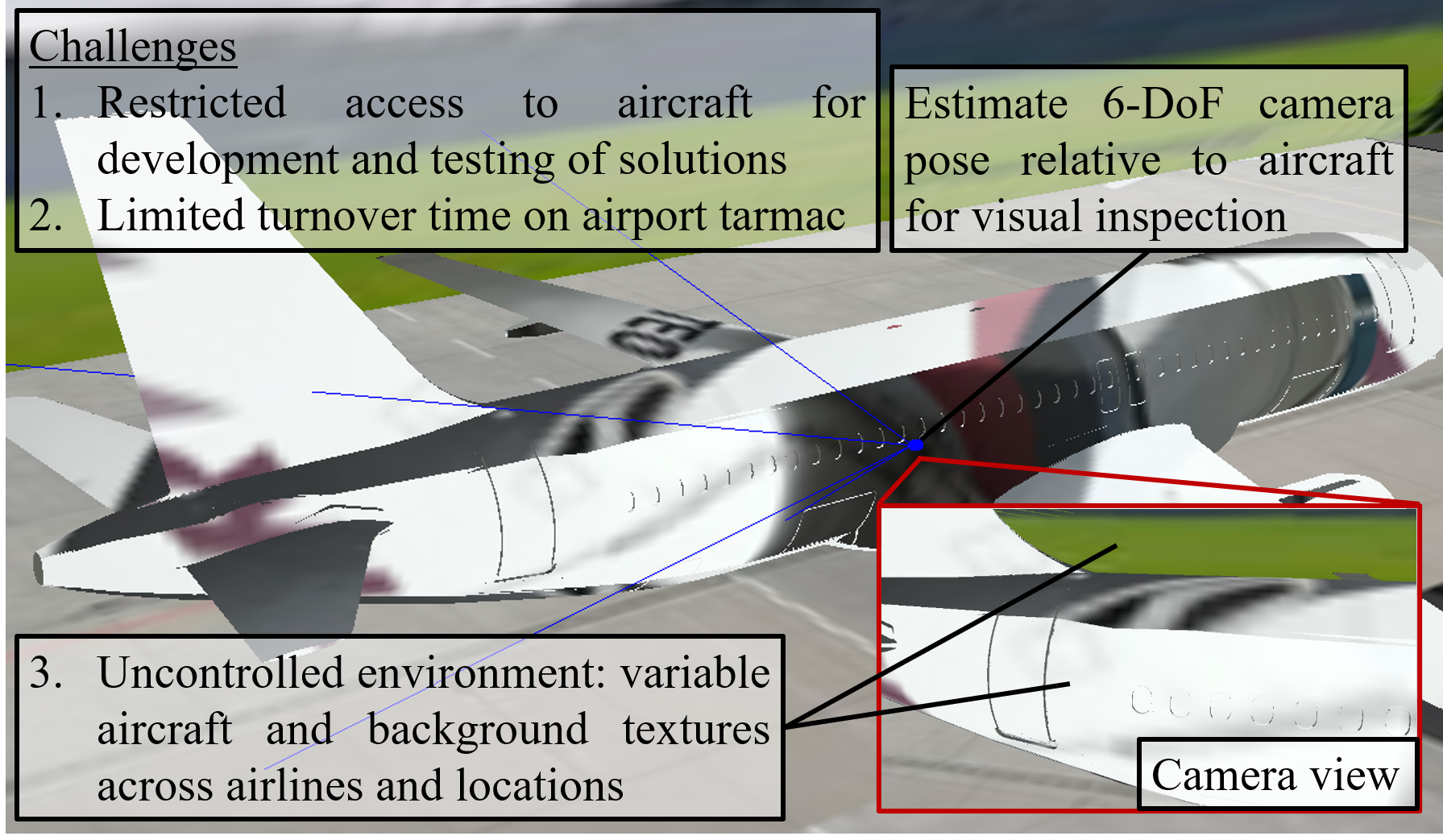}}
\vspace{-3mm}
\caption{Given the challenges of initialising an autonomous aircraft visual inspection system, we propose an infrastructure-less method to estimate a PTZ camera’s pose using a single image as input into a CNN.}
\label{fig:1}
\vspace{-6mm}
\end{figure}

Pan-Tilt-Zoom (PTZ) cameras are commonly used in inspection systems and a study in 2015 has explored the use of processing images taken from a PTZ camera on a mobile robot moving autonomously to pre-defined locations for aircraft exterior inspection \cite{R5}. However, it is only designed to inspect specific items (such as an oxygen bay handle or an air-inlet vent) of an aircraft and does not achieve precise localisation. Such solutions are also usually designed to be implemented in a hangar during maintenance, which provides a controlled environment to set up infrastructure for establishing checkpoints around the aircraft. Hence, these solutions are not suitable to be performed outdoors on the airport tarmac and with limited turnover time (approximately 2 hours) between flights.

In order to localise a detected defect, location information for each image (such as the coordinates of where each image centre coincides with the aircraft) has to be known and it is required to first determine the camera’s pose relative to the aircraft. Visual localisation methods such as Simultaneous Localisation and Mapping (SLAM) \cite{R6} and visual odometry \cite{R7} are capable of outdoor localisation, but they also require an initial pose estimation \cite{R8} to work. In addition, SLAM typically requires the payload to move around the aircraft for mapping, which is not suitable for the airport tarmac. Moreover, the highly restricted access to aircraft makes the development of new solutions challenging.

The problem of estimating a camera’s initial pose can be referred to as Camera Pose Estimation (CPE) and the recent increase in the use of Deep Convolutional Neural Networks (DCNNs) for monocular CPE has shown its potential to out-perform classical 3D structure-based methods in several aspects. The benefits of using DCNNs include shorter inference times, smaller memory \cite{R9}, and robustness to uncontrolled environments \cite{R10}. This shows the potential viability for the application of DCNNs for CPE as an infrastructure-less solution to outdoor aircraft inspection on airport tarmacs between flights, due to the time constraints and variations in lighting and background.

\restoregeometry

In this paper we propose an infrastructure-less method of estimating the pose of a PTZ camera with respect to an aircraft without the need for prior access to the real aircraft. This is achieved using a DCNN fine-tuned from pre-trained weights using only synthetic images obtained from a virtual camera capturing images of a 3D model of an aircraft. The PTZ camera is first roughly positioned and oriented within a reasonable boundary and faced perpendicular to the aircraft. The camera is then oriented to a fixed pan-tilt angle and an image is captured as input into a DCNN to regress a pose. We also leverage on known geometric information of the aircraft to modify an existing loss function and improve pose estimation accuracy. Our main contributions are as follows:

\begin{itemize}
\item The proposed method does not require infrastructure or prior access to the real aircraft during development. This is achieved using a pose regressor network trained on only synthetic images of a 3D aircraft model and a background applied with domain randomisation.
\end{itemize}
\begin{itemize}
\item We improved the network’s performance by introducing a new component to the network’s loss function. This additional loss component leverages on known geometry of an aircraft to provide a geometric relationship between the predicted position and orientation of the camera.
\end{itemize}

\section{Related Work}

CPE can be described as taking an input image and output an estimate of the pose – position and orientation – of the camera \cite{R9}. In most cases, the pose is obtained with respect to a predefined global reference. PoseNet \cite{R11} and other similar deep architectures that predicts a camera pose \cite{R12,R13,R14,R15,R16,R17,R18} share a common process, where images from a database are pre-processed before being used as input into a DCNN architecture for training. This training is conducted with the aim of minimising the error between the predicted pose and ground truth pose labels, represented by a loss function.

PoseNet \cite{R11} is the pioneer to introduce the use of a DCNN – based on a modified GoogLeNet \cite{R19} – to directly regress a camera pose, and many improved methods based on deep learning architectures have since been proposed \cite{R9}. These deep pose estimation methods have been fine-tuned and tested on publicly available indoor and outdoor datasets such as 7Scenes \cite{R20} and Cambridge Landmarks \cite{R11}. These datasets contain thousands of images, generated via sampling of videos recorded using hand-held camera devices such as a KinectFusion system \cite{R20} or a mobile phone before using software to automate the retrieval of camera poses of corresponding images \cite{R11}. However, fine-tuning models with these datasets may not accurately represent their effectiveness in CPE for our application due to the difficulties in obtaining a real dataset to train with in the first place, as well as the need to accommodate scene changes such as different appearance of the same aircraft model and its background.

Several researchers \cite{R21,R22,R23,R24,R25,R26} have explored fine-tuning DCNNs without the need for real images by using synthetic scenes for camera and object pose estimation tasks. Among these, \cite{R24} proposed a learning method for a drone to autonomously navigate an indoor environment without collision, by training a network through reinforcement learning using only 3D CAD models. Only RGB images rendered from a manually designed 3D indoor environment are used to train the CNN which outputs velocity commands. The authors apply random textures, object positions and lighting to create diverse scenes for their model to generalise and have managed to achieve autonomous drone navigation and obstacle avoidance in some indoor scenarios. While this explores the ability of a network trained using synthetic images to generalise to the real world, their objective was to avoid collision as opposed to CPE.

Recently, Acharya \textit{et al.} \cite{R26} have also proposed a solution for indoor CPE by fine-tuning PoseNet \cite{R11} using synthetic images rendered from a low-detail 3D indoor environment, modelled with reference to a Building Information Model (BIM). To avoid generating images for all possible positions and orientations within the 3D environment, the authors defined a boundary to the ground truths of the fine-tuning dataset. Images are captured by a virtual camera repositioned at 0.05m intervals along a pre-defined trajectory length of about 30m within the 3D building environment and kept within a height range of 1.5-1.8 m with a tilt of ${\pm} 10${\degree}. This work explores different methods of rendering, from cartoon-like to photo-realistic and textured to rendering only edges within each scene. They are able to estimate the camera’s position from real images with an accuracy of about 2 m. However, these existing methods have only been tested in known, controlled indoor environments with substantial changes in scenes and viewpoints as the camera relocates within the environment. This is as opposed to differentiating the camera pose between two images that are captured with slight changes in viewpoint in the context of aircraft GVI. Moreover, the reported accuracy in these works are insufficient for our application.

In a similar approach, Tobin \textit{et al.} \cite{R25} have investigated the use of domain randomisation to bridge the gap between simulation and reality. They argue that models fine-tuned with only synthetic scenes can generalise to real scenes if the scenes are diverse enough. The authors generate their dataset by uniformly randomising many aspects of their domain, including size, shape, position and colour of objects in each scene, and successfully taught a robotic arm to pick objects within a real, crowded indoor environment using only “low-fidelity” rendered images. Following this, others \cite{R27,R28,R29} have also applied domain randomisation for deep pose estimation tasks without training with real data. Despite being robust to object distractors, these models only apply to object pose estimation and often have other unchanging major objects such as a table where objects are placed on or a robot gripper which provides useful information of each object’s pose relative to these major objects in the scene. In our work, we focus on CPE with respect to an aircraft without any other known objects in the scene.

In summary, it can be very challenging to develop deep solutions for CPE with respect to an aircraft on an airport tarmac due to the limited access to real aircraft and solutions need to be robust to large variations in environment and aircraft texture across airlines. We propose to address this challenge by removing the need to obtain real images for training and use only synthetic images of the aircraft’s 3D model in scenes varied using domain randomization. Interestingly, recent works \cite{R9,R30} also suggest that deep pose regressors are out-performed by structure-based methods due to the lack of information on the scene’s geometry. In this paper, we leverage on known geometry of an aircraft’s surface to explore a geometric relationship between the camera's position and orientation within the network's loss function and improve the pose estimation accuracy.

\begin{figure}[t]
\centerline{\includegraphics[width=\columnwidth]{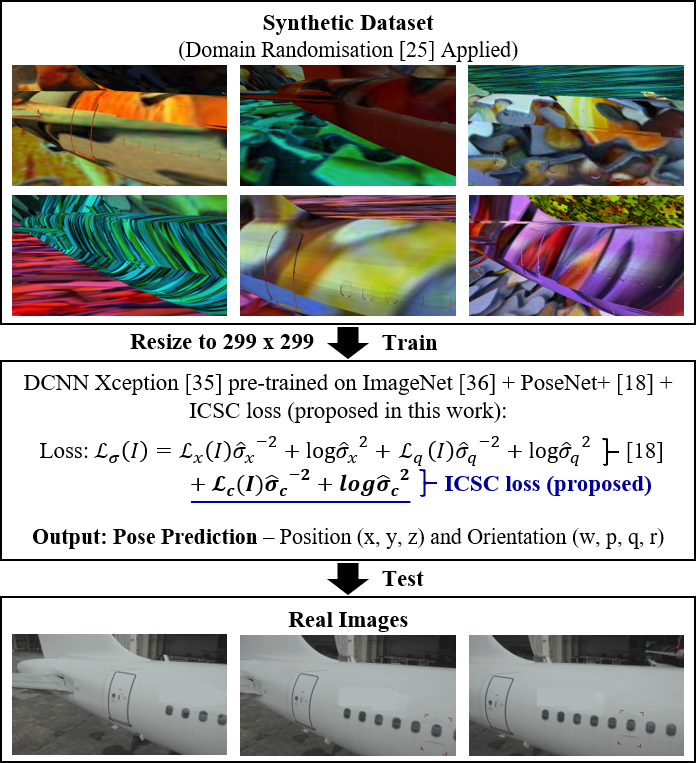}}
\vspace{-3mm}
\caption{Summary of deep learning approach.}
\label{fig:2}
\vspace{-5mm}
\end{figure}

\section{METHODOLOGY}
We propose a setup with realistic constraints and assumptions for how a PTZ camera of known specifications can be deployed next to an aircraft for the purpose of inspecting the upper surface of the aircraft’s fuselage. Based on this setup, we create a simple virtual 3D environment and capture images using a virtual camera while applying domain randomisation for our synthetic dataset. We use this synthetic dataset to train a deep network that can regress a camera’s pose from an input image. We base our network on an existing method, PoseNet with learnable weights \cite{R18} (we refer to as PoseNet+), in our approach and modify its loss function by introducing an additional loss component that provides a geometric relationship between the camera’s position and orientation. We summarise our deep learning approach in Fig. 2.

\subsection{Proposed Setup with PTZ Camera}

We use only the back half of an Airbus A320 (A320 in short) for illustration purpose. The following requirements and deployment steps are proposed:

\begin{figure}[t]
\centerline{\includegraphics[width=\columnwidth]{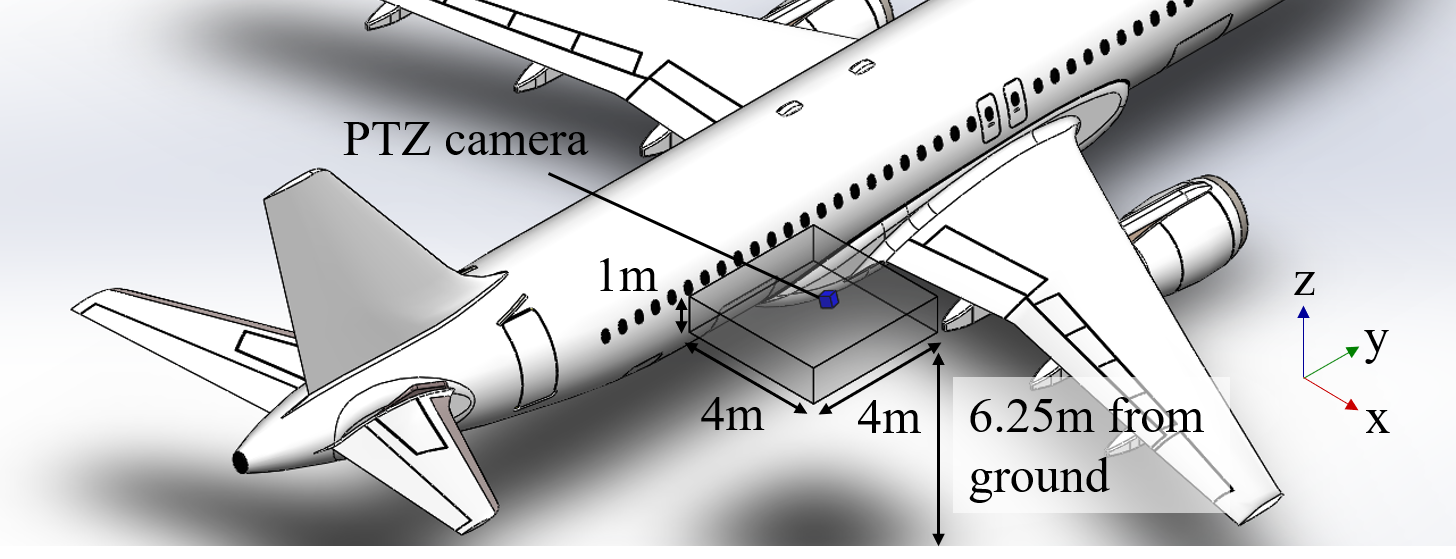}}
\vspace{-3mm}
\caption{Proposed boundary for PTZ camera position and axes direction.}
\label{fig:3}
\vspace{-4mm}
\end{figure}
\begin{figure}[t]
\centerline{\includegraphics[width=\columnwidth]{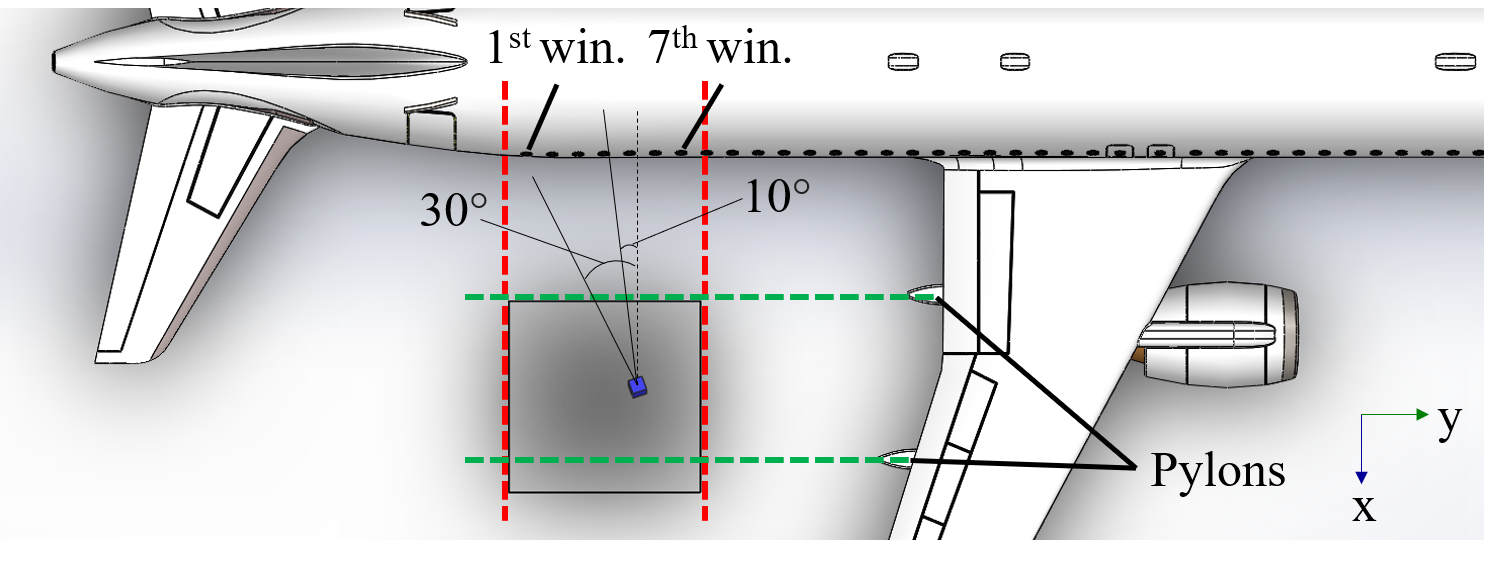}}
\vspace{-3mm}
\caption{Windows and pylons of an A320 are used as a visual guide to position the PTZ camera within the proposed boundary.}
\label{fig:4}
\vspace{-5mm}
\end{figure}

\begin{itemize}
\item The PTZ camera's specifications are known and the full Field of View (FOV) is used for initialisation.
\end{itemize}
\begin{itemize}
\item The PTZ camera can be positioned within a reasonable area of 4 m by 4 m and raised to a height of 6.25 m to 7.25 m from the ground via equipment such as an electronic mast or a boom lift and easily approximated with the use of accessible equipment such as a range finder.
\end{itemize}
\begin{itemize}
\item We assume that the PTZ camera’s base can be easily levelled (i.e. no roll and pitch relative to the ground) with the use of a gimbal or level gauge, and oriented about the z-axis to perpendicularly face the aircraft (within ${\pm} 10${\degree} yaw error). This reduces the problem to 4 Degrees of Freedom (DoF) - position and yaw.
\item The camera is then panned 20{\degree} towards the aircraft’s tail and tilted 18{\degree} towards the ground with commands sent via software and an image is captured for initialisation.
\end{itemize}

Fig.~\ref{fig:3} shows the proposed permissible space where the PTZ camera can be set up, while Fig.~\ref{fig:4} shows the features (windows and pylons) of an A320 to use as visual guides for the boundaries of this space. The yaw error of ${\pm} 10${\degree} due to manual orientation towards the aircraft suggests that the camera is oriented to between +10{\degree} to +30{\degree} about the z-axis.

\begin{figure}[t]
\centerline{\includegraphics[width=\columnwidth]{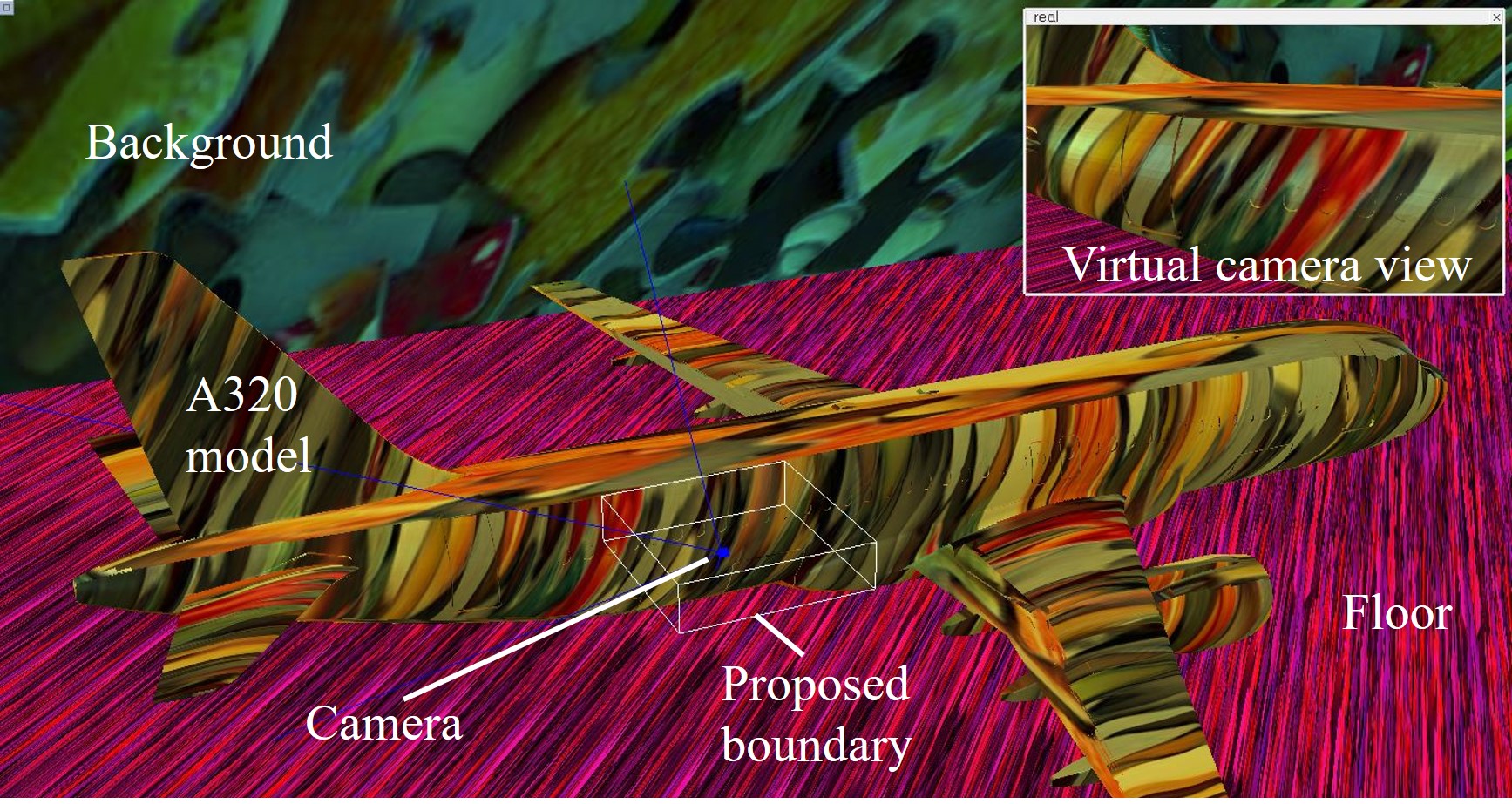}}
\vspace{-3mm}
\caption{Virtual setup with an instance of domain randomisation.}
\label{fig:5}
\vspace{-5mm}
\end{figure}

\subsection{Virtual Environment and Synthetic Dataset}
We have obtained the 3D model of an A320 from a GrabCAD contribution \cite{R31}. Using SolidWorks \cite{R32}, minor modifications are made to match general features and overall dimensions of a real A320, based on details obtained from an A320’s Structural Repair Manual (SRM). 

Our virtual setup is shown in Fig. 5. To create this 3D environment, we place our 3D model into a scene in robot simulator CoppeliaSim \cite{R33}. A large wall is added on one side of the aircraft model as background. A virtual camera is placed beside the aircraft and its FOV is set to match the real PTZ camera (61.6{\degree} horizontal FOV at 16:9 aspect ratio).

We apply domain randomisation \cite{R25} when generating our synthetic dataset as it has been demonstrated to be capable of generalising to real-world data, given sufficient simulated variability. We use a free stock image of a randomly scattered puzzle to apply as texture for the ground, aircraft model, and background. We randomise the following aspects when capturing each image for our dataset:

\begin{itemize}
\item The PTZ camera’s position within the proposed 4 m x 4 m x 1 m boundary;
\item The PTZ camera’s pan between +10{\degree} to +30{\degree}, rotated about the z-axis;
\item The PTZ camera’s tilt between -17.5{\degree} to -18.5{\degree}. (slight tolerance of ${\pm} 0.5${\degree} from the proposed 18{\degree} tilt;
\item Colour – RGB values of both ambient and specular components for the texture of every object; and
\item The position, orientation, as well as horizontal and vertical scaling factors of textures applied onto each surface.
\end{itemize}
A total of 4700 synthetic images are generated, of which 4000 are used for training and 700 for validation.

\subsection{Base Deep Learning Architecture}

PoseNet \cite{R11} was introduced in 2015 and is first to use a DCNN to directly regress for CPE. Its model is based on modifying GoogLeNet’s (Inception v1) \cite{R19} architecture, which was then considered as the state-of-the-art DCNN for image classification. PoseNet made modifications to the model, including rescaling each input image and performing a centre crop to match GoogLeNet’s 224x224 pixel input and adding a fully connected layer while replacing all softmax layers (used for classification) with regression layers that output both position (x, y, z) and orientation vectors (quaternions – w, p, q, r). PoseNet also redefine its loss function as:

\begin{equation}
\mathcal{L}_\beta(I) =  \mathcal{L}_x(I) + \beta\mathcal{L}_q(I)
\label{eq:1}
\end{equation}
where \begin{math}\mathcal{L}_x(I) = \|x-\hat{x}\|_2\end{math} and \begin{math}\mathcal{L}_q(I) = \|q-\frac{\hat{q}}{\|\hat{q}\|}\|_2\end{math}, with \begin{math}\hat{x}\end{math} and \begin{math}\hat{q}\end{math} representing the predicted position and orientation vectors respectively, while \begin{math}x\end{math} and \begin{math}q\end{math} represent ground truth pose. In practice, it is observed that \begin{math}q\end{math} is close enough to \begin{math}\hat{q}\end{math} and the normalization of \begin{math}\hat{q}\end{math} is removed from the loss function during implementation.  \begin{math}\beta\end{math} is a hyperparameter that functions as a factor to scale the orientation error in attempt to keep the position and orientation errors similar.

However, substantial effort is required when tuning \begin{math}\beta\end{math} to obtain a reasonable balance between the orientation and position losses. To address this, PoseNet+ \cite{R18} proposes a loss function that learns a weighting between the position and orientation components. The loss function is formulated using the concept of homoscedastic uncertainty - a measure of uncertainty of the task and is independent of input data \cite{R34} - and is defined as:
\begin{equation}
\mathcal{L}_\sigma(I) =  \mathcal{L}_x(I)\hat{\sigma}^{-2}_x + \log\hat{\sigma}^2_x + \mathcal{L}_q(I)\hat{\sigma}^{-2}_q + \log\hat{\sigma}^2_q
\label{eq:2}
\end{equation}where \begin{math}\hat{\sigma}^2_x\end{math} and \begin{math}\hat{\sigma}^2_q\end{math} represent the homoscedastic uncertainties and are optimised with respect to the loss function through back propagation. While the variance \begin{math}\sigma^{2}\end{math} is learnt, the logarithmic regularisation term prevents the network from learning an infinite variance to achieve zero loss. During implementation, \begin{math}\hat{s}:=\log\hat{\sigma}^2\end{math} is learnt as it avoids a potential division by zero and the function becomes:
\begin{equation}
\mathcal{L}_\sigma(I) =  \mathcal{L}_x(I)\exp(-\hat{s}_x) + \hat{s}_x + \mathcal{L}_q(I)\exp(-\hat{s}_q) + \hat{s}_q
\label{eq:3}
\end{equation}

We base our method on loss function \eqref{eq:3} as it has been proven to substantially outperform PoseNet's original loss function \eqref{eq:1}. We apply this method onto a more recent deep architecture, Xception \cite{R35} (improved from Inception v3), as it results in substantially better performance than GoogLeNet (Inception v1). We made slight modifications to the Xception network in a similar fashion to PoseNet, by replacing the softmax layer with a regression layer with 7 pose outputs. We resize every input image to match the network's 299 x 299 pixel input size without a centre crop as we found this to improve performance and attribute this to the increase in features and other spatial information that may be present in the whole image despite the distortion from resizing.

\subsection{Image Centre Scene Coordinate (ICSC) Loss}

We propose to modify loss function \eqref{eq:3} by introducing an additional component \begin{math}c\end{math} that uses the scene coordinate of each image’s centre pixel. This is obtained by finding the point of intersection between the equation describing the camera’s viewpoint (as a function of \begin{math}x\end{math} and \begin{math}q\end{math}) and the aircraft's surface. With our proposed setup, we find this point of intersection is always on the upper half of the fuselage and propose to model the aircraft's surface as the equation of a cylinder. With \begin{math}c\end{math} as the cartesian coordinates of any point on the cylinder's surface, the equation of the surface is given by:
\begin{equation}
c_x^2 + (c_z-h_0)^2 = r_0^2
\label{eq:4}
\end{equation}Where \begin{math}c_x\end{math}, \begin{math}c_y\end{math} (any value along the cylinder’s length) and \begin{math}c_z\end{math} are the coordinates of a point on the cylinder's surface, \begin{math}h_0\end{math} is the displacement of the cylinder’s cross-sectional centre from the scene’s origin, and \begin{math}r_0\end{math} is the aircraft’s fuselage radius. The line representing the camera’s viewpoint is formulated as:
\begin{equation}
\vec{l} = \vec{x} + t\vec{v}
\label{eq:5}
\end{equation}Where \begin{math}\vec{l}\end{math} represents the camera’s viewpoint, \begin{math}\vec{x}\end{math} is the camera's position, \begin{math}\vec{v}\end{math} is obtained by rotating the camera’s default direction vector by quaternion \begin{math}q\end{math}, and \begin{math}t\end{math} is a variable that determines the position of any point along line \begin{math}\vec{l}\end{math}.

For every pair of camera position and orientation, we use equations \eqref{eq:4} and \eqref{eq:5} to solve for \begin{math}t\end{math} where \begin{math}c = \vec{l}\end{math} to determine the point of intersection between line \begin{math}\vec{l}\end{math} and the surface of the cylinder. Since a line may intersect the surface of a cylinder at up to two points, only the point nearest to the camera’s position, \begin{math}x\end{math}, is kept. Fig.~\ref{fig:6} illustrates how the aircraft fuselage’s surface is modelled as the surface of a cylinder, as well as how \begin{math}\mathcal{L}_x\end{math} and \begin{math}\mathcal{L}_q\end{math} can be related by \begin{math}\mathcal{L}_c\end{math}. We combine our proposed loss component with \eqref{eq:2} to result in:

\begin{figure}[t]
\centerline{\includegraphics[width=\columnwidth]{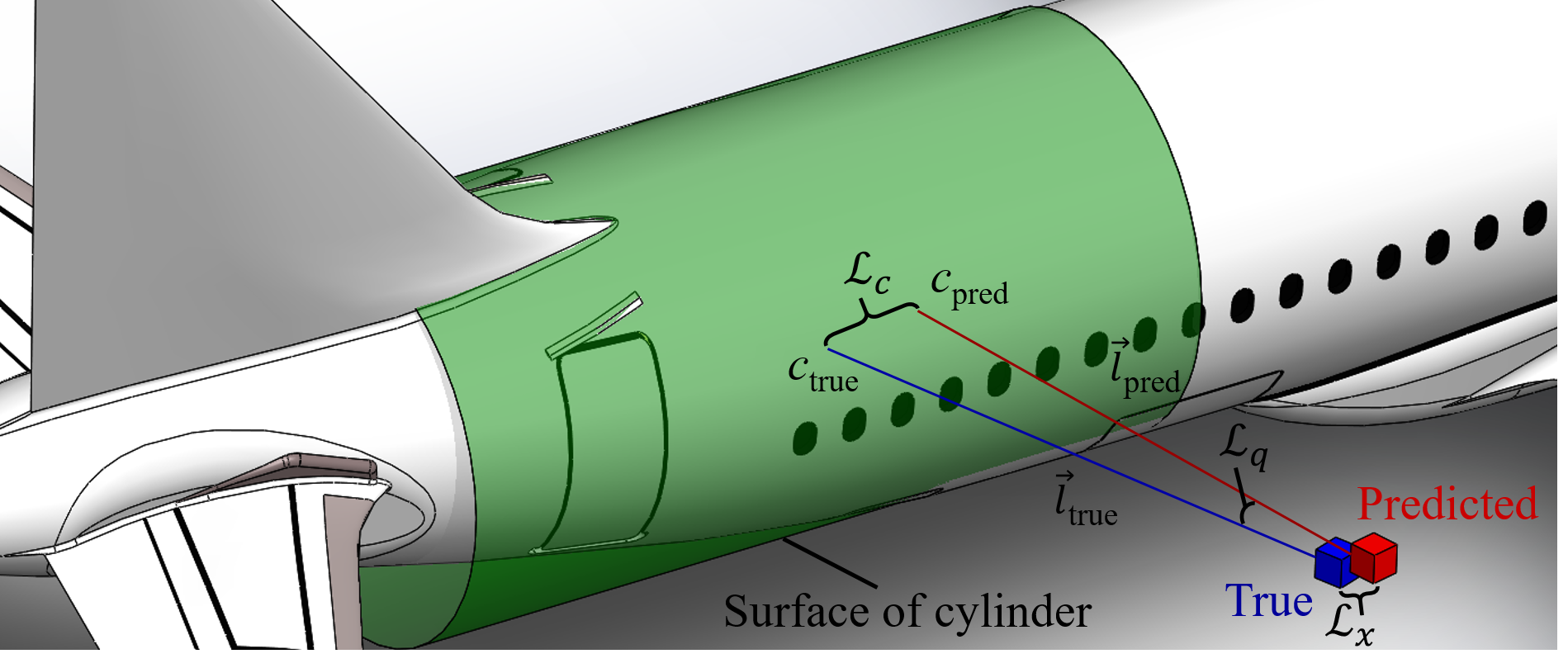}}
\vspace{-3mm}
\caption{Visualisation of how the aircraft’s surface is modelled as the surface of a cylinder (green). Losses \begin{math}\mathcal{L}_x\end{math}, \begin{math}\mathcal{L}_q\end{math} and \begin{math}\mathcal{L}_c\end{math} can also be visualised as a difference between their respective true and predicted components.}
\label{fig:6}
\vspace{-3mm}
\end{figure}

\begin{multline}
\mathcal{L}_\sigma(I) =  \mathcal{L}_x(I)\hat{\sigma}^{-2}_x +\log\hat{\sigma}^2_x + \mathcal{L}_q(I)\hat{\sigma}^{-2}_q
+\log\hat{\sigma}^2_q \\ + \mathcal{L}_c(I)\hat{\sigma}^{-2}_c +\log\hat{\sigma}^2_c
\label{eq:6}
\end{multline}
Where \begin{math}\mathcal{L}_c(I) = \|c-\hat{c}\|_2\end{math}, and \begin{math}c-\hat{c}\end{math} represents the difference between the true and predicted point of intersection coordinates. The following function is then implemented:
\begin{multline}
\mathcal{L}_\sigma(I) =  \mathcal{L}_x(I)\exp(-\hat{s}_x) + \hat{s}_x + \mathcal{L}_q(I)\exp(-\hat{s}_q) + \hat{s}_q \\
+ \mathcal{L}_c(I)\exp(-\hat{s}_c) + \hat{s}_c
\label{eq:7}
\end{multline}Where \begin{math}\hat{s}_x\end{math}, \begin{math}\hat{s}_q\end{math} and \begin{math}\hat{s}_c\end{math} are learnt and we arbitrarily initialise all of them to zero. We refer to our proposed additional loss component as the Image Centre Scene Coordinate (ICSC) loss and the network with loss function \eqref{eq:7} as ICSC-PoseNet.

\section{Experiments and Results}

\subsection{Obtaining Real Images with Ground Truths}
\begin{figure}[t]
\centerline{\includegraphics[width=\columnwidth]{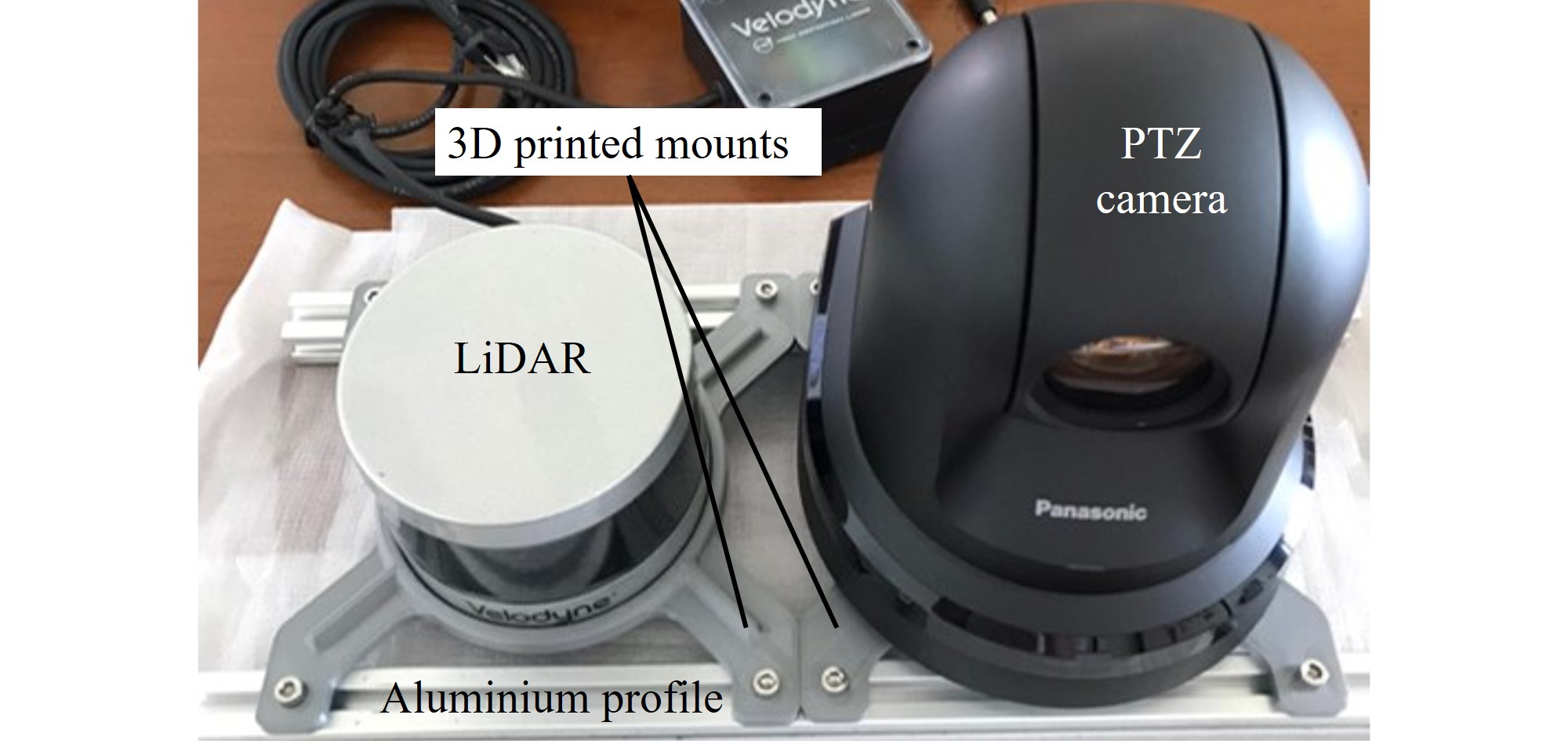}}
\vspace{-3mm}
\caption{A prototype rig: a LiDAR and a PTZ camera mounted onto their respective 3D printed mounts and secured onto a pair of aluminium extrusions.}
\label{fig:7}
\vspace{-3mm}
\end{figure}

To evaluate our approach, we have requested and obtained special access to an A320 in an outdoor area to obtain images and pose-related data for our real test dataset. We build a prototype consisting of a Panasonic AW-HE40H PTZ Camera and a Velodyne VLP-16 3D LiDAR secured onto a rig as shown in Fig.~\ref{fig:7}. The 3D LiDAR is used to obtain ground truth and is not used in the proposed methodology. While we can analyse the point cloud from the 3D LiDAR to obtain the camera pose, the process is very troublesome and time consuming. Multiple adjustments are required to ensure the desired features are present within the point cloud to obtain each acceptable pair of image and point cloud.

The prototype is powered by a portable power bank and programmatically accessed from a laptop. The prototype is brought within the proposed boundary atop a boom lift’s platform. Images for this setup are not shown due as they are deemed sensitive by the venue and airline. A total of 28 images with ground truths are obtained as our real dataset for testing. For comparison of the coverage, our training images are generated within +5 m to +9 m, -9.25 m to -5.25 m, +6.25 m to +7.25 m, and +10{\degree} to +30{\degree} in the x, y, z and pan respectively, while the ground truth of real test images spans +7 m to +8.5 m, -9.2 m to -6.3 m, +6.9 m to +7.2 m, and +11{\degree} to +26{\degree} in the x, y, z and pan respectively.

\begin{figure*}[t]
    \centering
    
  \subfloat[Real images of aircraft used as input.\label{1a}]{%
       \includegraphics[width=\textwidth]{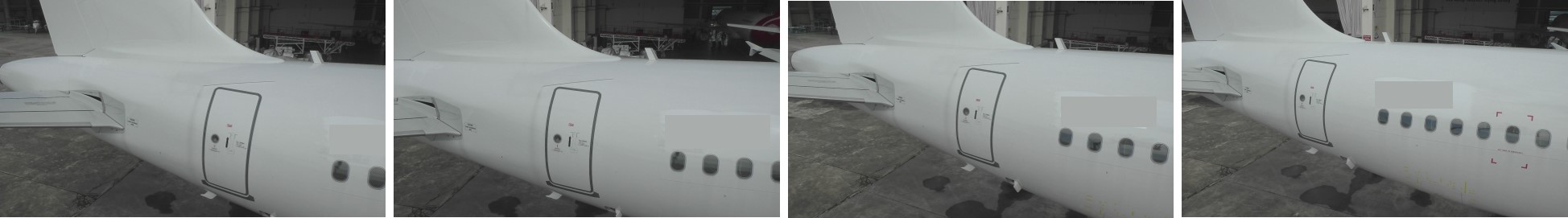}}
\vspace{-2mm}

  \subfloat[Predictions by PoseNet+ \cite{R18}.\label{1b}]{%
        \includegraphics[width=\textwidth]{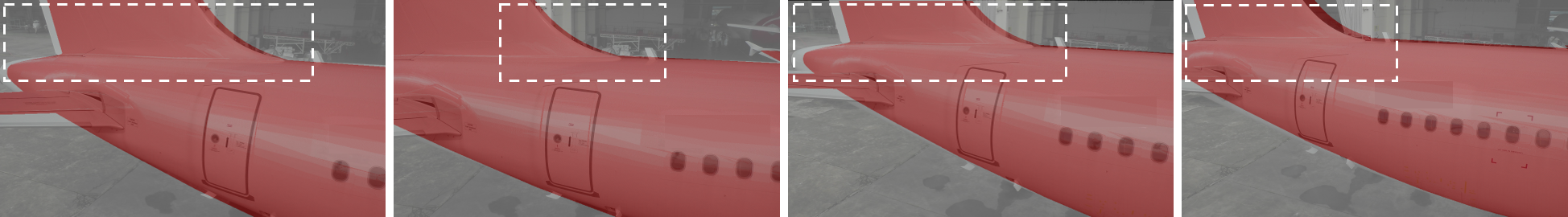}}
\vspace{-2mm}

  \subfloat[Predictions by ICSC-PoseNet (proposed method) results in slightly better overlap of the aircraft.\label{1c}]{%
        \includegraphics[width=\textwidth]{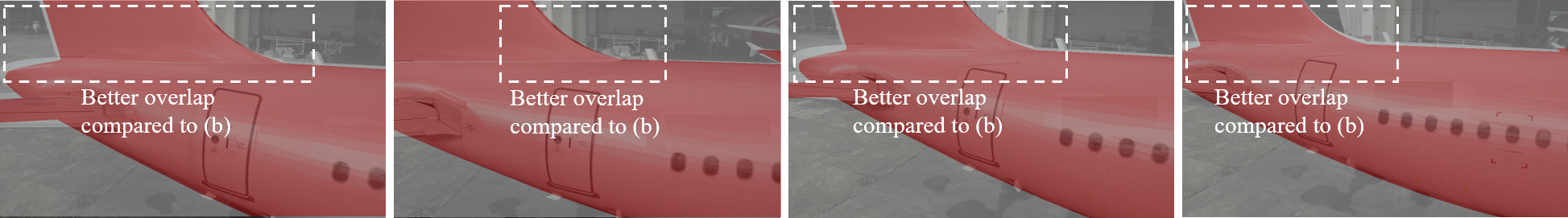}}

\setlength\abovecaptionskip{0.3\baselineskip}
  \caption{From the dataset of 28 real test images, we choose four images (a) that return the largest prediction error by both networks. (b) and (c) show virtual images captured from the camera pose predicted by the two networks, overlaid in red on the real input images (a). Comparing (b) and (c) qualitatively, we can see that predictions by ICSC-PoseNet shows better overlap of the aircraft as compared to the predictions by PoseNet+ \cite{R18} (b).}
  \label{fig:8} 
  \vspace{-5mm}
\end{figure*}

\subsection{Implementation and Results}
Our network is implemented using TensorFlow, supported by a NVIDA RTX Turbo 2080Ti GPU. For clarity, we refer to ‘training’ as fine-tuning our network pre-trained on ImageNet \cite{R36} to leverage on transfer learning. We also normalise all input images such that all pixel intensities range from -1 to 1. We optimise our network with ADAM \cite{R37} using default parameters at a learn rate of \begin{math}10^{-4}\end{math} and a batch size of 25.

We train two networks with Xception \cite{R35} as their base architecture, one as PoseNet+ with the learnable weighting loss function \eqref{eq:3} as the baseline and the other as ICSC-PoseNet with the modified loss function \eqref{eq:7}. The networks are evaluated by testing on the 28 real images with ground truths. We train each network for 200 epochs as we observe that overfitting tends to occur beyond that. Table~\ref{tab1} provides the best results as well as a comparison with PoseNet+. Four of the 28 real images that return the largest error when predicted by both networks are shown in Fig.~\ref{fig:8}, along with their prediction results overlaid in red on the real images.

\section{Discussion}

\subsection{Deploying a PTZ Camera Within Proposed Boundaries}
The spatial coverage of the real images obtained with their ground truths demonstrates that the proposed method of using aircraft features (windows and pylons) as landmarks to guide the positioning and orienting of the PTZ camera within our boundary is feasible. This is an important step since the use of DCNN for CPE performs best within a pre-defined range of predictions (for both position and orientation) that should be included in the training dataset. This limitation has been discussed in literature, where deep pose estimators still underperform in the task of generalising to unseen scenes \cite{R9} and perform more similar to image retrieval methods \cite{R30}. We show that we can manually position our PTZ camera within the same proposed boundary and pan range used to generate our synthetic training images.

\newcolumntype{C}{>{\centering\arraybackslash}X}
\newcolumntype{R}{>{\raggedright\arraybackslash}X}

\begin{table}[t]
\caption{Results - Position and Orientation Error}
\vspace{-3mm}
\begin{center}
\begin{tabularx}{\columnwidth}{|R|C|C|C|}
\hline
Loss & \multicolumn{3}{c|}{{Error (lowest error in bold)}} \\
\cline{2-4} 
Function & Median & MAE$^{\mathrm{a}}$ & RMSE$^{\mathrm{b}}$ \\
\hline
PoseNet+[18] (Learnable Weighting) & 0.292m, 1.252{\degree} & 0.303m, 1.278{\degree} & 0.312m, 1.437{\degree} \\
\hline
ICSC-PoseNet (this work) & \textbf{0.217m, 0.731{\degree}} & \textbf{0.226m, 0.815{\degree}} & \textbf{0.237m, 0.882{\degree}} \\
\hline
\multicolumn{4}{l}{$^{\mathrm{a}}$Mean Absolute Error, $^{\mathrm{b}}$Root-Mean-Square Error.}
\vspace{-8mm}
\end{tabularx}
\end{center}
\label{tab1}
\end{table}

\subsection{Camera Pose Estimation Without Training on Real Images}
Our results demonstrate the network’s ability to estimate a PTZ camera’s pose within a region next to an Airbus A320 without training on any real images. This is achieved without any knowledge of the scene other than the aircraft model, and without any infrastructure. Using single images as input into our network, we obtain a median prediction error of 0.217m and 0.731{\degree} (Table~\ref{tab1}) which is sufficient for initialisation given the scale of the aircraft. For comparison, the window-to-window distance of an A320 is about 0.53m. In Fig.~\ref{fig:8}, prediction results overlaid in red over four real images show the network’s ability to extract relevant aircraft features from the randomised textures in the synthetic training dataset and match their scale and position with the real images to regress a pose. We observe a high degree of overlap even in the four samples with the largest prediction error (up to 0.49m and 3.06{\degree}) across all predictions by both networks.

\subsection{Comparison of Loss Functions}
We quantitatively compare results for pose prediction in Table~\ref{tab1} and find that training the network with our additional ICSC loss component in ICSC-PoseNet substantially improves pose prediction accuracy. Lower error is observed across Median Error, Mean Absolute Error (MAE) and Root-Mean-Square Error (RMSE), in both position and orientation predictions by ICSC-PoseNet as compared to PoseNet+. Qualitatively, we also observe in Fig.~\ref{fig:8} a slight improvement in the region of overlap of the aircraft in the images from predictions by our network (ICSC-PoseNet) as compared to the predictions by PoseNet+. While the improvement seems minor visually, it can substantially impact the accuracy of defect localisation on the aircraft when the camera is zoomed in during inspection. We conclude that the addition of a component in the loss function that geometrically relates the position and orientation predictions during training can improve camera pose estimation accuracy in our application.

\section{Conclusion}
We have demonstrated that camera pose estimation with respect to an aircraft can be achieved without any infrastructure or prior access to a real aircraft. This is achieved through the proposed ICSC-PoseNet, which successfully reduced pose estimation error by leveraging on geometric information of an aircraft to introduce an additional component to the loss function of an existing network. In the future, we plan to adapt the method to perform sensor fusion with other sensor data such as from a LiDAR to improve performance.

\section*{Acknowledgment}
This research is supported by ST Engineering Aerospace as part of a project with the Civil Aviation Authority of Singapore to develop a GVI system for detecting damage to the exterior of aircraft due to lightning strikes.

\bibliographystyle{./bibliography/IEEEtran}
\bibliography{./bibliography/main}

\end{document}